# A Proof-of-Concept Study of Artificial Intelligence–Assisted Contour Revision

Ti Bai[1], Anjali Balagopal[1], Michael Dohopolski[1], Howard E. Morgan[1], Rafe McBeth[1], Jun Tan[1], Mu-Han Lin[1], David J. Sher[1], Dan Nguyen[1], and Steve Jiang[1*]

[1]Medical Artificial Intelligence and Automation (MAIA) Laboratory, Department of Radiation Oncology, University of Texas Southwestern Medical Center, Dallas, Texas 75390

*Corresponding Author: Steve.Jiang@UTSouthwestern.edu

ABSTRACT

**Background and purpose:** Automatic segmentation of anatomical structures is critical for many medical applications. However, the results are not always clinically acceptable and require tedious manual revision. Here, we present a novel concept called artificial intelligence–assisted contour revision (AIACR) and demonstrate its feasibility.

**Materials and Methods:** The proposed clinical workflow of AIACR is as follows: given an initial contour that requires a clinician's revision, the clinician indicates where a large revision is needed, and a trained deep learning (DL) model takes this input to update the contour. This process repeats until a clinically acceptable contour is achieved. The DL model is designed to minimize the clinician's input at each iteration and to minimize the number of iterations needed to reach acceptance. In this proof-of-concept study, we demonstrated the concept on 2D axial images of three head-and-neck cancer datasets, with the clinician's input at each iteration being one mouse click on the desired location of the contour segment. The performance of the model is quantified with Dice Similarity Coefficient (DSC) and 95$^{th}$ percentile of Hausdorff Distance (HD95).

**Results:** The average DSC/HD95 (mm) of the auto-generated initial contours were 0.82/4.3, 0.73/5.6 and 0.67/11.4 for three datasets, which were improved to 0.91/2.1, 0.86/2.4 and 0.86/4.7 with three mouse clicks, respectively. Each DL-based contour update requires ~20 ms.

**Conclusion:** We proposed a novel AIACR concept that uses DL models to assist clinicians in revising contours in an efficient and effective way, and we demonstrated its feasibility by using 2D axial CT images from three head-and-neck cancer datasets.

## 1  INTRODUCTION

Radiotherapy plays an important role in cancer treatment. Its success relies on elaborate treatment planning to deliver the prescription radiation dose to the planning target volume (PTV) while appropriately sparing organs-at-risk (OARs) from unnecessary radiation. The OARs and PTV are delineated upstream of the planning process during the segmentation step. Since the segmentation result affects all downstream processes, accurate OAR and target contours are essential.

Since manual contouring can be labor intensive and time consuming, automatic segmentation of OARs and target volumes is highly desirable, especially for online adaptive radiotherapy. Tremendous efforts have been devoted to meeting this important clinical need for timely contour delineation, such as the traditional graph cut algorithms [1], the well-known atlas-based segmentation algorithms [2], and the popular registration-driven auto-segmentation algorithms [3]. Deep learning (DL)-based models represent the current state of the art [4-7]. We direct readers to this review article [8] for a comprehensive summary of these DL-based medical image segmentation algorithms.

However, given the imperfections of current automatic segmentation algorithms, clinicians must inspect and manually revise automatically generated contours for incident errors. This manual revision process can be very time consuming, sometimes requiring a comparable amount of time to manually contouring from scratch [9], which heavily hampers the clinical implementation of these auto-segmentation models.

As part of our ongoing efforts to develop Artificial Intelligence and Clinician Integrated Systems (AICIS), we aim to alleviate this practical problem and to facilitate the clinical implementation of auto-segmentation models by using AI as an assistant to clinicians. In this work, we propose a novel concept of using AI to assist clinicians in revising contours, which we have termed Artificial Intelligence–Assisted Contour Revision (AIACR), and we present a proof-of-concept study to demonstrate its feasibility. We envision the following workflow for AIACR: given an input medical image, an initial contour is generated via an auto-segmentation model (or generated by a trainee); this initial contour is then reviewed, and revised when needed, by the clinician. During the revision process, the clinician indicates where the current contour should be with a simple mouse click at the desired contour boundary. The AI model updates the contour based on this input from the clinician, and the process repeats until an acceptable contour is achieved.

As an assistant to the clinician, the goal of the AIACR model is to minimize the clinician's inputs at each iteration and the number of iterations required. In this work, at each iteration, we employ one mouse click on the desired boundary as the clinician's input for efficient and controllable contour revision. We assume that the clinician will preferentially revise the contour segment with the largest errors, since this would require fewer iterations/clicks.

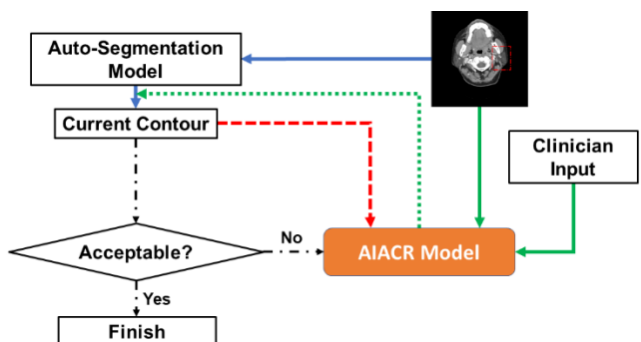

Figure 1: The clinical workflow based on the proposed AIACR model for AI-assisted contour revision. First, an auto-segmentation model generates the initial contour (pipeline indicated by the blue solid lines), which the clinician reviews (pipeline indicated by the black dot-dash lines). If it is not acceptable, the current contour (as indicated by the red dash line) and the image, as well as the clinician input (as indicated by the green solid lines), are fed into the AIACR model to update the contour (as indicated by the green dot line). This process repeats until an acceptable contour is achieved.

## 2 METHODS AND MATERIALS

### A. Methods

The clinical workflow based on the proposed AIACR model is illustrated in Figure 1. An initial contour is generated by an auto-segmentation model (or by a trainee) and reviewed by the clinician. If the contour is clinically acceptable, then the contouring process is finished. Otherwise, given the current contour, the clinician provides one mouse click on the desired boundary in the area with a large error (i.e., where the contour should be). This mouse click–based feedback guides the AIACR model to update the contour.

Specifically, the mouse click is converted into a 2D image (hereafter termed the "click image") by placing a 2D Gaussian point with a radius of 10 pixels around the above boundary point. The input of our AIACR model consists of three channels: the original image, the current segmentation mask, and the click image. The output is the updated contour.

The clinician then reviews the updated contour. If it is acceptable, then the revision process is finished. Otherwise, the clinician makes a second mouse click in the area currently exhibiting the next largest error (again, where the contour should be). In this case, both the first and the second mouse clicks are converted into a single joint 2D click image by placing two 2D Gaussian points (each with a radius of 10 pixels) around the above two clicked boundary points. The updated click image is fed into the AIACR model to further revise the contour. This process is repeated until the resultant contour meets the clinician's expectations.

*B.* Network Architecture

In this work, we employed the basic U-Net [6] architecture as our AIACR model for proof-of-concept purposes.

To be specific, in the encoder part, continuous stride-two convolutional layers are applied to extract the image's high-level features. The feature number is doubled per each downsampling operation by using stride-two convolutional operators until it reaches a feature number of 512. The downsampling depth is set as eight, such that the bottleneck layer has a feature size of $1 \times 1$. In the decoder part, a concatenation operator is adopted to fuse the low-level and the high-level features. With regard to the network modules, each convolutional layer contains three consecutive operators: convolution, instance normalization [10] and rectified linear unit (ReLU). The convolutional operator has a kernel size of $3 \times 3$. The initial feature number in the input layers is 64. The input and output channels are three and one, respectively.

*C.* Datasets

*1)* Training and validation datasets

We used the open access head-and-neck (HN) CT scans of patients with nasopharyngeal cancer from the Automatic Structure Segmentation for Radiotherapy Planning Challenge to demonstrate the algorithm. Each CT scan was marked by one experienced oncologist and verified by another experienced oncologist. We randomly split the datasets into 40 and 10 patients, which served as the training and validation datasets, respectively. Twenty-one annotated OARs were employed for the performance evaluation.

All of the original CT scans in this dataset consist of 100~144 slices of $512 \times 512$ pixels, with a voxel resolution of ($[0.98$~$1.18] \times [0.98$~$1.18] \times 3.00$ mm$^3$). Since we used the 2D slice CT images to demonstrate the algorithm, we had 4941/1210 images for training/validation, respectively.

*2)* Testing datasets

It is important to demonstrate a model's generalizability by using data with previously unseen demographics and distributions. In this work, we employed two testing datasets distinct from the training and validation datasets.

The first testing dataset was released by DeepMind. This dataset consists of de-identified HN CT scans that were initially segmented with full volumetric regions by a radiographer with at least four years of experience, then arbitrated by a second radiographer with similar experience. A radiation oncologist with at least five years of post-certification experience arbitrated further still. Each scan in this dataset has 21 annotated OARs and consists of 119~184 slices of $512 \times 512$ pixels, with a voxel resolution of ($[0.94$~$1.25] \times [0.94$~$1.25] \times 2.50$ mm$^3$). In this work, we used 28 scans consisting of 4427 2D slice CT images to test the model's performance. Ten of the 21 OARs overlapped with those used in the training dataset, so we used them to quantify the performance.

The second testing dataset was collected at UT Southwestern Medical Center (UTSW) and approved by the IRB. This dataset contains 20 patient scans. The contours of the OARs were exported from the clinical system and cleaned with our in-house software tools, followed by a manual double-check. Each scan has at most 53 annotated OARs. Depending on clinical tasks, different patients may have different OARs annotated. All the scans in this dataset contain 124~203 slices of $512 \times 512$ pixels, with a voxel resolution of ([1.17~1.37] × [1.17~1.37] × 3.00 mm³). In total, we have 2980 2D slice CT images. Ten OARs overlapped with the above training dataset and were thus used to evaluate the algorithm.

Due to considerations of computational efficiency, for each volumetric CT scan in the training, validation and testing datasets, we cropped out a sub-volume with an axial size of 256 × 256 so that the clinically significant HN regions were covered.

### D. Training Details

We used both Dice loss and Hausdorff distance (HD)-based loss [11] for model training. We used a weight to balance these two losses such that the weighted HD-based loss had values similar to the Dice loss for any training sample. We used Adam optimizer [12] with parameters $\beta_1 = 0.9$ and $\beta_2 = 0.999$ to update the model parameters at every $1 \times 10^5$ iterations. The learning rate was initially set as $1 \times 10^{-4}$, which was then reduced to $1 \times 10^{-5}$ and $1 \times 10^{-6}$ at iterations $5 \times 10^4$ and $7.5 \times 10^4$, respectively. The batch size was one.

It should be noted that there are no interactive segmentation datasets with real clinicians' feedback signals. Indeed, it is very hard, if not impossible, to collect such data. Therefore, in this work, we purposely constructed an interactive segmentation dataset in an online fashion during the training phase by simulating the clinician's mouse click with the highest chance to be the desired boundary point with the largest distance from the current contour.

In detail, given the currently predicted contour (denoted as $C_p$), we first calculated the distance of each point on the ground truth contour $C_g$ to the current predicted contour $C_p$ as below:

$$D_{C_p}(y) = \inf_{x \in C_p} d(x, y) \; \forall y \in C_g \tag{1}$$

where $d(x, y)$ represents the distance between the points on $C_p$ (denoted as $x$) and the points on $C_g$ (denoted as $y$). In other words, the above function, $D_{C_p}(y)$, represents the distance for every point on the ground truth contour to the nearest point on the predicted contour. Assuming that the clinicians will likely click on the boundary points that correspond to the large errors, we converted this distance measurement to click probability to use during the training phase. We used the SoftMax transformation to assign a click probability to each boundary point on the desired (ground truth) contour as follows:

$$P(y) = \frac{exp\left(-D_{C_p}(y)\right)}{\sum_{y' \in C_g} exp\left(-D_{C_p}(y')\right)} \; \forall y \in C_g \tag{2}$$

Now, given the current segmentation map, we could randomly sample the clinician's click point on the desired contour based on the probabilistic distribution defined by Equation (2). The larger the error that a boundary point on the desired contour corresponded to, the higher the chance that this point would be sampled (clicked by the clinician).

Given the randomly sampled point, the current segmentation map and the input CT image, we constructed three-channel input data to feed into the AIACR model for training. The supervised signal was the ground truth segmentation mask.

In this study, we used an in-house DL-based auto-segmentation model to generate the initial contours for all the patient cases. To illustrate the performance of AIACR on the poorly segmented initial contours, which is what we propose to use AIACR for, we purposely left the auto-segmentation model unoptimized.

*E.* Testing details and evaluation metrics

To quantify the performance of our AIACR model, during the testing phase, without loss of generality, we simulated the clinician's mouse clicking at each iteration by choosing the boundary point corresponding to the largest error, i.e., the largest distance defined by Equation (1), which is the one-side HD from the ground truth contour $c_g$ to the currently predicted contour $c_p$.

We used two different metrics to quantify the model performance: Dice coefficient (DSC) and the 95$^{th}$ percentile of the HD (HD95).

Nine of the twenty-one OARs annotated in the validation dataset were deemed clinically relevant or more challenging for auto-segmentation: brainstem, spinal cord, pituitary, left and right parotid glands, left and right inner ears, and left and right mandibles. Four of those OARs were included in the DeepMind and UTSW testing datasets: brainstem, spinal cord, and left and right parotid glands. Accordingly, we used these representative OARs (nine in the validation set, four in the other two sets) as showcases to illustrate the gradual improvement in model performance from 1~3 clicks.

For an overall quantification, we computed the average values of the above two metrics among all the investigated OARs separately in the validation, DeepMind and UTSW testing datasets. We also report the model inference efficiency, i.e., the time the model takes to produce a new revised contour.

3   RESULTS

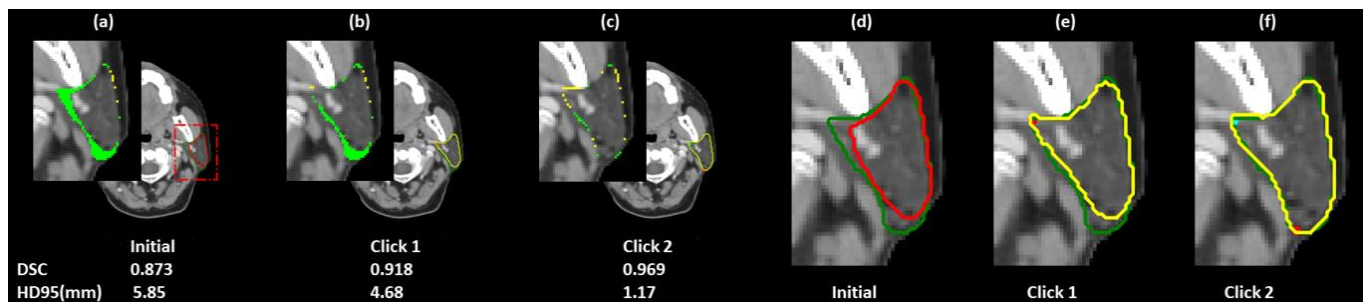

Figure 2: Showcase for the segmentation of a left parotid gland from the validation dataset with 2 clicks. (a)-(c) show the segmentation results for the following time points: initial (a), after 1 click (b), and after 2 clicks (c) with AIACR. Associated segmentation errors are demonstrated in the top left corner of each cell (green: false negative, yellow: false positive). Note how the area occupied by the segmentation errors decreases with each successive click, thus demonstrating the effectiveness of AIACR with user input. (d)-(f) show the zoomed-in contour comparisons of the left parotid glands (green: ground truth contour, red: initial contour, yellow: updated contour at each iteration) along with the clicked points (red: current clicks, cyan: past clicks). Display window: [-160, 240].

Figure 2 shows an example of using AIACR to contour a left parotid gland case in the validation dataset. Figure 2(d) clearly reveals a large surface gap between the ground truth contour (green) and the initial contour (red). After one click (red dot in Figure 2[e]), the section of the contour around the clicked point improved significantly. The associated DSC/HD95 values improved from 0.873/5.85 mm (initial segmentation) to 0.918/4.68 mm (first iteration). After the second click (red dot in Figure 2[f]), the updated contour closely approximated the ground truth contour, with DSC/HD95 values of 0.969/1.17 mm. This progressive improvement with each mouse click can also be observed in the segmentation error maps, where substantial false negative errors were corrected. From the segmentation error map after the second click (in Figure 2[c]), we can see that the difference between the predicted contour and the ground truth contour is about one pixel.

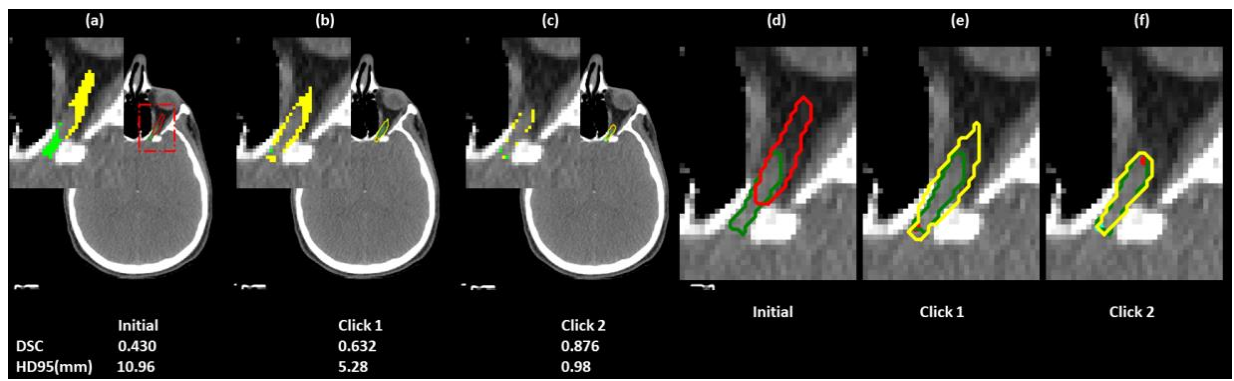

Figure 3: Showcase for the segmentation of a left optic nerve from the DeepMind test dataset with 2 clicks. (a)-(c) show the segmentation results for the following time points: initial (a), after 1 click (b), and after 2 clicks (c) with AIACR. Associated segmentation errors are demonstrated in the top left corner of each cell (green: false negative, yellow: false positive). Note how the area occupied by the segmentation errors decreases with each successive click, thus demonstrating the effectiveness of AIACR with user input. (d)-(f) show the zoomed-in contour comparisons (green: ground truth contour, red: initial contour, yellow: updated contour at each iteration) along with the clicked points (red: current clicks, cyan: past clicks). Display window: [-160, 240].

Figure 3 shows an example of a left optic nerve from the DeepMind test dataset. Again, Figure 3(f) shows that a close approximation between the revised contour (yellow) and the ground truth contour (green) can be achieved after two mouse clicks from the clinician. The error map (Figure 3[c]) shows that the differences between the revised segmentation mask and the ground truth mask are within one pixel. By contrast, the initial segmentation error map (Figure 3[a]) displays large false positive and false negative results even though the optic nerve exhibits a clear boundary. Quantitatively, after two clicks, the DSC/HD95 metrics improved dramatically from 0.43/10.96 mm to 0.876/0.98 mm, respectively.

Figure 4 shows an example of a brainstem case from the UTSW test dataset. We can see that each click corrected one segment around the clicked point. The initial segmentation result shows segmentation errors, especially in the anterior direction.

Figure 5 illustrates the gradual improvement in performance based on selected organs in three different datasets. We can see that the DSC/HD95 increases/decreases substantially for all the testing cases with each click (1-3).

Table 1 provides a further quantitative demonstration of the average performance improvement based on all the organ cases in the three different datasets. It reveals a more than 10 percent absolute increase in DSC on all three datasets after three clicks, while HD95 was cut almost in half. Moreover, the poorer the initial performance (UTSW dataset), the greater the improvement that AIACR achieved.

The time required to update the contour at each iteration was ~20 ms when using a single NVIDIA GeForce Titan X graphics card, which would allow the clinician and the AI to interact in real time during the contour revision process.

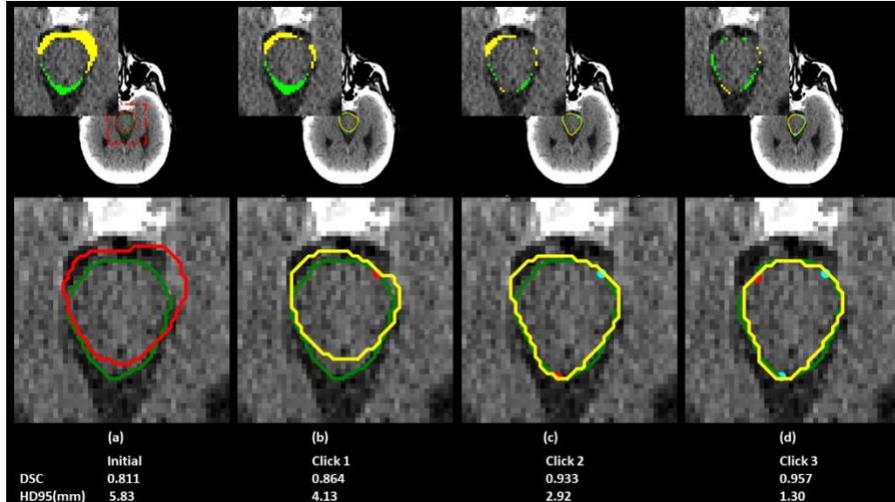

Figure 4: Showcase for the segmentation of a brainstem from the UTSW dataset that achieved excellent accuracy with 3 clicks. (a)-(d) show the segmentation results for the following time points: initial (a), after 1 click (b), after 2 clicks (c), and after 3 clicks (d) with AIACR. Associated segmentation errors are demonstrated in the top left corner (green: false negative, yellow: false positive). Note how the area occupied by the segmentation errors decreases with each successive click, thus demonstrating the effectiveness of AIACR with user input. The bottom row shows the zoomed-in contour comparisons (green: ground truth contour, red: initial contour, yellow: updated contour at each iteration) along with the clicked points (red: current clicks, cyan: past clicks). Display window: [0, 80].

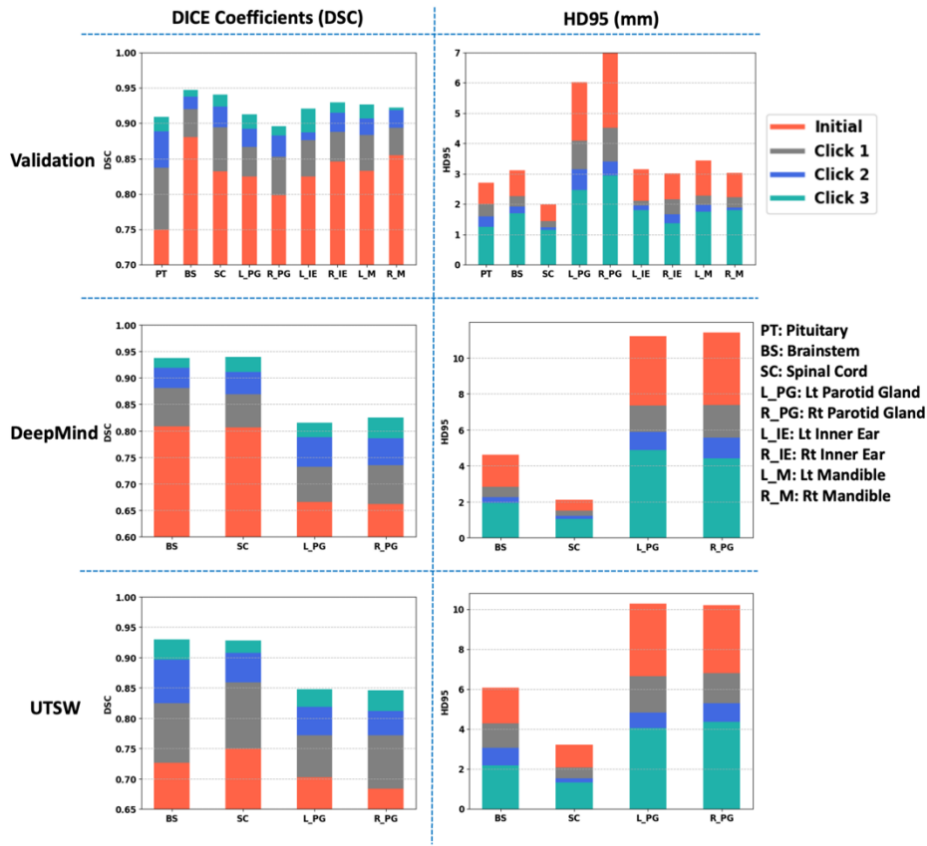

Figure 5: Gradual performance improvement with each click based on selected clinically significant and relatively challenging organ cases. The Dice coefficients (left) and HD95 (right) are used to quantify the performance improvement for three datasets: validation (top), DeepMind (middle), and UTSW (bottom). The x-axis represents different organs, whose full names are listed on the right.

| DSC/HD95(mm) | Validation | DeepMind | UTSW |
|---|---|---|---|
| Initial | 0.82/4.3 | 0.73/5.6 | 0.67/11.4 |
| Click 1 | 0.87/3.0 | 0.78/3.6 | 0.76/7. 5 |
| Click 2 | 0.89/2.4 | 0.83/2.8 | 0.82/5.7 |
| Click 3 | 0.91/2.1 | 0.86/2.4 | 0.86/4.7 |

Table 1: Performance improvement with three clicks quantified by the averaged Dice Coefficients (DSC) and HD95 among all the organs for different datasets.

4 DISCUSSIONS AND CONCLUSIONS

This work is part of our ongoing effort towards developing AI- and clinician-integrated systems (AICIS) in medicine. AICIS directly addresses a problem that most AI models face when deployed into clinical practice. The current dogma for implementing AI clinically typically involves AI and clinicians working independently and sequentially: AI independently performs a clinical task, such as segmentation, and presents the outcome to clinicians to review, then they accept, reject, or revise the outcome. In contrast, AICIS attempts to integrate AI into clinicians' existing workflows, allowing them to perform clinical tasks collaboratively to achieve a clinically acceptable result in a more efficient and user-friendly manner; this will facilitate clinicians' acceptance of AI being implemented into routine clinical practice.

For the clinical task of organ and tumor segmentation, specifically, after the auto-segmentation model generates the initial contour, the clinician may face three options: accept as is (AAI), accept with revision (AWR), and reject (REJ). We always try to improve the accuracy of auto-segmentation models to maximize the AAI ratio and minimize the REJ ratio. If there were a perfect auto-segmentation model that produced a 100% AAI ratio, AI could independently finish the clinical task without clinicians' inputs, which would potentially allow AI to replace clinicians. In the real world, no model can achieve a 100% AAI ratio, and for especially challenging cases, clinicians' manual revisions of the initial contours are always needed. Accordingly, we propose AIACR to assist clinicians in revising the initial contours in an efficient and user-friendly way. AIACR is also not intended to replace auto-segmentation models. Instead, it works downstream of an auto-segmentation model, so it can be used with any state-of-the-art auto-segmentation model.

One of the advantages of our AIACR model is that it can alleviate the generalizability issue from which many DL-based auto-segmentation models suffer. Even if an auto-segmentation model generalizes poorly to outside datasets, using AIACR downstream can greatly improve these results, as shown by the major improvement in the average performance from the initial to the revised contours for the UTSW and DeepMind testing datasets (Table 1).

AIACR will likely play an important role in online adaptive radiotherapy (ART). The current pipeline for ART includes acquiring a same day image (e.g., cone beam CT [CBCT]) and deforming or creating new OARs and target structures to optimize the radiation treatment plan, given the patient's current anatomy. This approach can account for gas passing through the intestines or tumors shrinking from current therapy. However, a significant barrier to implementing online ART is the time required to manually correct contours. Therefore, improving the efficiency of clinicians' contour revision could improve the acceptance of online ART by radiotherapy departments.

Several online ART modalities are currently being investigated clinically, including CBCT-based [13], MRI-based [14], and PET/CT-based techniques [15]. Using current CBCT-based adaptive approaches for head-and-neck cancer as an example, one study evaluated the contouring accuracy and efficiency of the online ART process in the Ethos™ emulator environment [9]. The authors reported that the median time spent in this virtual online ART process was almost 20 minutes (range 13 to 31 minutes), where the most time was spent reviewing and editing contours [9]. If AIACR could reduce the time spent reviewing and editing contours, then this adaptive process could be more easily integrated into the already strained time requirements of a busy practice. Therefore, evaluating AIACR in the adaptive setting is of interest for future experiments.

For simplicity in this proof-of-concept study, we employed 2D axial images to demonstrate the feasibility of AIACR. In the future, we will extend this novel concept to 3D cases and conduct a comprehensive, clinically realistic evaluation.

In summary, we have proven the concept of AIACR. Experimental results show that, with the aid of AIACR, DSC/HD95 values improved dramatically from 0.82/4.3 to 0.91/2.1 (validation), 0.73/5.6 to 0.86/2.4 (DeepMind), and 0.67/11.4 to 0.86/4.7 (UTSW) with just three clicks. Furthermore, given the swiftness of the model response time, with each contour update only requiring about 20 ms, AIACR can work in real time.

DATA ACCESS

The training and validation dataset are openly accessible at https://structseg2019.grand-challenge.org/Dataset/. The DeepMind dataset are also openly accessible at https://github.com/deepmind/tcia-ct-scan-dataset. The UTSW dataset is currently not publicly available for ethical reasons but is available from the corresponding author upon reasonable request and if approved by the IRB.

ACKNOWLEDGMENTS

We would like to thank Dr. Jonathan Feinberg for editing the manuscript. The data in the UTSW dataset were collected at UT Southwestern Medical Center under an approved IRB protocol.